\pdfoutput=1

\documentclass[11pt]{article}
\usepackage{authblk}
\usepackage[]{ACL2023}

\usepackage{times}
\usepackage{latexsym}

\usepackage[T1]{fontenc}

\usepackage[utf8]{inputenc}

\usepackage{microtype}

\usepackage{inconsolata}
\usepackage{microtype}
\usepackage{amsmath, amssymb, amsthm, amscd, amsfonts}
\usepackage{IEEEtrantools}
\usepackage{times}
\usepackage{soul}
\usepackage{url}
\usepackage[utf8]{inputenc}
\usepackage{graphicx}
\usepackage{amsmath}
\usepackage{amsthm}
\usepackage{booktabs}
\usepackage{algorithm}
\usepackage{algorithmic}
\usepackage{natbib} 
\usepackage{amssymb}
\usepackage{amsfonts} 
\usepackage{multirow}
\usepackage{boldline}
\usepackage{hhline}
\usepackage{amsmath}
\usepackage{xcolor}
\usepackage{pifont}
\usepackage{subfigure}
\usepackage{makecell}
\usepackage{mathrsfs}
\usepackage{utfsym}
\usepackage[normalem]{ulem}
\useunder{\uline}{\ul}{}
\usepackage[inline]{enumitem}
\usepackage{enumitem}
\usepackage{IEEEtrantools}
\usepackage{stmaryrd}
\usepackage{mathabx}

\theoremstyle{definition}

\theoremstyle{remark}

\title{A Confidence-based Partial Label Learning Model for \\Crowd-Annotated Named Entity Recognition}

\author{
    \bf{\normalsize
    Limao Xiong$^{1}$, \ \ Jie Zhou$^{1}$\thanks{$^*$  Corresponding author, jie\_zhou@fudan.edu.cn.} , \ \ Qunxi Zhu$^{2}$, \ \ Xiao Wang$^{1}$, \ \ Yuanbin Wu$^{3}$, \ \ Qi Zhang$^{1}$,} \\ 
    \bf{\normalsize Tao Gui$^{4}$, \ \ Xuanjing Huang$^{1}$, \ \ Jin Ma$^{5}$, \ \ Ying Shan$^{5}$} \\
  {$^1$ \normalsize School of Computer Science, Fudan Univerisity} \\
  {$^2$ \normalsize Research Institute of Intelligent Complex Systems, Fudan University} \\
  {$^3$ \normalsize The Department of Computer Science and Technology, East China Normal University} \\
  {$^4$ \normalsize Institute of Modern Languages and Linguistics, Fudan University} \\
  {$^5$ \normalsize Applied Research Center (ARC), Tencent PCG} \\
  }

\begin{document}
\maketitle

\renewcommand{\thefootnote}{\fnsymbol{footnote}}
\renewcommand{\thefootnote}{\arabic{footnote}}

\begin{abstract}
Existing models for named entity recognition (NER) are mainly based on large-scale labeled datasets, which always obtain using crowdsourcing. 
However, it is hard to obtain a unified and correct label via majority voting from multiple annotators for NER due to the large labeling space and complexity of this task.  
To address this problem, we aim to utilize the original multi-annotator labels directly. 
Particularly, we propose a Confidence-based Partial Label Learning (\texttt{CPLL}) method to integrate the prior confidence (given by annotators) and posterior confidences (learned by models) for crowd-annotated NER. 
This model learns a token- and content-dependent confidence via an Expectation–Maximization (EM) algorithm by minimizing empirical risk. 
The true posterior estimator and confidence estimator perform iteratively to update the true posterior and confidence respectively. 
We conduct extensive experimental results on both real-world and synthetic datasets, which show that our model can improve performance effectively compared with strong baselines.
\end{abstract}

\section{Introduction}
Named entity recognition (NER) plays a fundamental role in many downstream natural language processing (NLP) tasks, such as relation extraction \cite{bach2007review}, event extraction \cite{wadden2019entity,zhou2022multi}. Recently, by leveraging deep learning models, existing NER systems have witnessed superior performances on NER datasets.
However, these models typically require a massive amount of labeled training data, such as MSRA \cite{levow2006third}, Ontonotes 4.0 \cite{weischedel2011ontonotes}, and Resume \cite{zhang2018chinese}. 
In real applications, we often need to consider new types of entities in new domains where we do not have existing annotated.
The majority way to label the data at a lower cost is crowdsourcing \cite{peng2015named}, which labels the data using multiple annotators. 

The crowd-annotated datasets are always low quality for the following two reasons.
First, as an exchange, crowd annotations are always nonexperts. Various annotators may have different interpretations of labeling guidelines. Moreover, they may make mistakes in the labeling process. It is hard to require a number of annotations to reach an agreement.
For example, annotator 1 labels ``David and Jack" as a PER entity, while the correct label is ``David" and ``Jack" under our guidelines (Table \ref{fig:example}). Also we should label the continuous time and place as one entity (e.g, ``tomorrow at 10:00 a.m." and ``company ( room 1003 )"). 
Second, due to the ambiguous word boundaries and complex composition, the NER task is more challenging compared with the text classification tasks. 
Annotator 3 ignores the token ``a.m." for the time entity and adds ``the" as part of the place entity falsely. Also, he/she misses the person entities in the text.
In this paper, we focus on building a powerful NER system based on crowd-annotated data, which is of low quality. 

There are two main ways to utilize crowd-annotated data. 
One simple and important way to obtain high-quality annotations for each input instance is majority voting. 
As shown in Table \ref{fig:example}, the majority voting method can not obtain the correct answers from these three annotations well. The right labels (e.g., ``David", ``Jack", ``tomorrow at 10:00 a.m.", and ``company ( room 1003 )") are only annotated by annotators 1 and 2 once.
Another majority of work models the differences among annotators by finding the trustworthy annotators \cite{rodrigues2014sequence,nguyen2017aggregating,yang2018adversarial}. 
From Table \ref{fig:example}, we can find that none of the three annotators labels the entities absolutely right. 
Thus, these two kinds of methods are a waste of human labor.

\begin{table*}[h!]
\centering
\small
\begin{tabular}{ll}
\hlineB{3}
 Annotator 1 &  $\overbrace{\mathrm{\colorbox{green!30}{David~and~Jack}}}^{\mathrm{PER~\textcolor{red}{\usym{2717}}}}$ will hold a meeting $\overbrace{\mathrm{\colorbox{blue!30}{tomorrow~at~10:00~a.m.}}}^{\mathrm{TIME~\textcolor{green}{\usym{2713}}}}$ in the $\overbrace{\mathrm{\colorbox{red!30}{company~(~room~1003~)}}}^{\mathrm{PLACE~\textcolor{green}{\usym{2713}}}}$.\\ \hline
 Annotator 2 &  $\overbrace{\mathrm{\colorbox{green!30}{David}}}^{\mathrm{PER~\textcolor{green}{\usym{2713}}}}$ and $\overbrace{\mathrm{\colorbox{green!30}{Jack}}}^{\mathrm{PER~\textcolor{green}{\usym{2713}}}}$ will hold a meeting $\overbrace{\mathrm{\colorbox{blue!30}{tomorrow}}}^{\mathrm{TIME~\textcolor{red}{\usym{2717}}}}$ at $\overbrace{\mathrm{\colorbox{blue!30}{10:00~a.m.}}}^{\mathrm{TIME~\textcolor{red}{\usym{2717}}}}$ in the $\overbrace{\mathrm{\colorbox{red!30}{company}}}^{\mathrm{PLACE~\textcolor{red}{\usym{2717}}}}$ ( $\overbrace{\mathrm{\colorbox{red!30}{room~1003}}}^{\mathrm{PLACE~\textcolor{red}{\usym{2717}}}}$ ).\\\hline
 Annotator 3 &  David and Jack will hold a meeting $\overbrace{\mathrm{\colorbox{blue!30}{tomorrow~at~10:00}}}^{\mathrm{TIME~\textcolor{red}{\usym{2717}}}}$ a.m. in $\overbrace{\mathrm{\colorbox{red!30}{the~company~(~room~1003}}}^{\mathrm{PLACE~\textcolor{red}{\usym{2717}}}}$ ). \\
\hlineB{3}
\end{tabular}
\caption{The spans marked with \colorbox{blue!30}{blue}, \colorbox{green!30}{green}, and \colorbox{red!30}{red} are time (TIME), person (PER), and place (PLACE) entities labeled by three annotators.}
\label{fig:example}
\end{table*}

To address this problem, we translated this task into a partial label learning (PLL) problem, which trains the model based on the dataset where each sample is assigned with a set of candidate labels \cite{cour2011learning,wen2021leveraged}. 
Thus, it is natural to utilize all human labor via PLL, which can be divided into two types: 1) average-based methods which consider each candidate class equally \cite{hullermeier2006learning,zhang2015solving}; 2) identification-based methods which predict the ground-truth label as a latent variable via a translation matrix to describe the scores of each candidate label \cite{feng2019partial,yan2020partial,feng2020provably}. 
Despite extensive studies on PLL methods, there are still two challenges in our condition.
One challenge (\textbf{C1}) is that these methods are criticized when the same candidate label occurs more than once.
The general PLL is under the assumption that each candidate label is only been assigned once, while each sample may be assigned the same classes multiple times by the different annotators in our situation.
Another challenge (\textbf{C2}) is that most of the existing studies about PLL focus on image or text classification tasks, while we focus on a more complex task, sequence labeling, where each token is asserted with a label. Thus, the token itself and its content should be considered in this task. 

In this paper, we propose a Confidence-based Partial Label Learning (\texttt{CPLL}) model for crowd-annotated NER. 
For \textbf{C1}, we treat the classes' labeled number for each sample as prior confidence provided by the annotators. 
Also, we learn the confidence scores via an Expectation–Maximization (EM) algorithm \cite{dempster1977maximum}. 
We estimate the real conditional probability $P(Y=y|T=t, X=\mathbf{x})$ via a true posterior estimator based on the confidence that consists of the prior and posterior confidences. 
For \textbf{C2}, we learn a token- and content-dependent confidence via a confidence estimator to consider both the token $t$ and sequence input $\mathbf{x}$, because the candidate labels are always token-dependent and content-dependent.
In fact, our model can be applied to all the sequence labeling tasks, such as word segment, part of speech, etc.
We conduct a series of experiments on one real-world dataset and four synthetic datasets. The empirical results show that our model can make use of the crowd-annotated data effectively. We also explore the influence of annotation inconsistency and balance of prior and posterior confidences.

The main contributions of this work are listed as follows.
\begin{itemize}[leftmargin=*, align=left]
    \item To better utilize the crowd-annotated data, we propose a \texttt{CPLL} algorithm to incorporate the prior and posterior confidences for sequence labeling task (i.e., NER).
    \item To take the confidence scores into account, we design a true posterior estimator and confidence estimator to update the probability distribution of ground truth and token- and content-dependent confidence iteratively via the EM algorithm. 
    \item Extensive experiments on both real-world and synthetic datasets show that our \texttt{CPLL} model outperforms the state-of-the-art baselines, which indicates that our model disambiguates the noise labels effectively.
\end{itemize}

\begin{figure*}[t!]
\begin{center}
\includegraphics[width=1.0\textwidth]{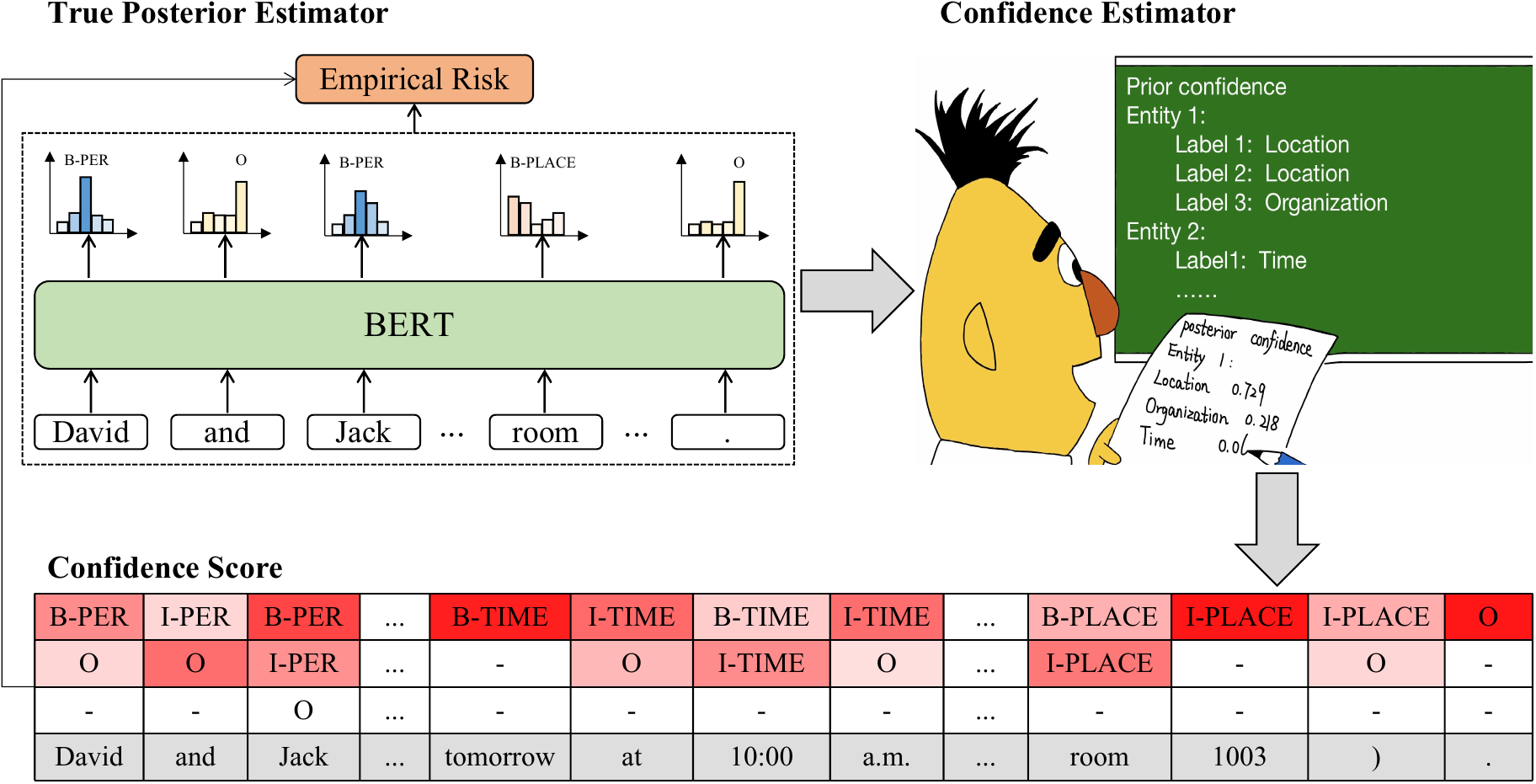}
\end{center}
\caption{The framework of our \texttt{CPLL} model, which consists of a true posterior estimator and confidence estimator. The true posterior estimator is used to predict the true posterior $P(Y=y|T=t, X=\mathbf{x})$ based on the confidence score learned by the confidence estimator. The confidence estimator learns the confidence based on the prior confidence obtained from annotators and the posterior confidence learned by the model.} 
\label{fig:framework}
\vspace{-2mm}
\end{figure*}

\section{Our Approach}
In this section, we first give the formal definition of our task. Then, we provide an overview of our proposed \texttt{CPLL} model. Finally, we introduce the main components contained in our model.

\subsection{Formal Definition}
Given a training corpus $\mathcal{D}=\{\mathbf{x}_i, (\hat{Y}_i, A_i)\}_{i=1}^{|\mathcal{D}|}$, where $\mathbf{x}=\{t_1, t_2, ..., t_{|\mathbf{x}|}\}$, ($\hat{Y}, A)=\{(\hat{\mathbf{y}}_1, \mathbf{a}_1), (\hat{\mathbf{y}}_2, \mathbf{a}_2), ..., (\hat{\mathbf{y}}_{|\mathbf{x}|}, \mathbf{a}_{|\mathbf{x}|})\}$.
Here, $\hat{\mathbf{y}}=\{y_1, y_2, ...., y_{|\hat{\mathbf{y}|}}\}$ is the candidate label set of the token $t$ and $\mathbf{a}=[a_1, a_2, ..., a_{|\hat{\mathbf{y}}|}]$ is labeled times obtained from annotations. 
Specifically, $a$ is the labeled times of candidate label $y$ for token $t$. 
$\hat{\mathbf{y}} \in \{2^{\mathcal{Y}} \setminus \emptyset \setminus \mathcal{Y} \}$ where $\mathcal{Y}$ is the label space and $2^{\mathcal{Y}}$ means the power set. 
For the rest of this paper, $y$ denotes the true label of token $t$ in text $\mathbf{x}$ unless otherwise specified. 
The goal of this task is to predict the truth posterior probability $P(Y=y|T=t, X=\mathbf{x})$ of token $t$ in text $\mathbf{x}$.

\subsection{Overview}
In this paper, we propose a CONfidence-based partial Label Learning (\texttt{CPLL}) model for crowd-annotated NER (Figure \ref{fig:framework}). 
Particularly, we learn the true posterior $P(Y=y|T=t, X=\mathbf{x})$ via a true posterior estimator $f$ and a confidence score $g(y; \hat{Y}, t, \mathbf{x})$ by minimizing the following risk. 
\begin{equation}
\small
\label{eq1}
\begin{aligned}
    R = \mathbb{E}_{p(\mathbf{x}, \hat{\mathbf{y}})}\left[\sum_{t \in \mathbf{x}}\sum_{y}\underbrace{g(y; \hat{\mathbf{y}}, t, \mathbf{x})}_{\mathrm{Confidence}} * \underbrace{\mathcal{L}(f(y; t, \mathbf{x}), y)}_{\mathrm{True~posterior}}\right]
\end{aligned}
\end{equation}
where the classifier $f(y; t, \mathbf{x})$ is used to predict $P(Y=y|T=t, X=\mathbf{x})$ and $\mathcal{L}$ is the loss. 
Particularly, we rely on the Expectation-Maximization algorithm \cite{dempster1977maximum} to find the maximum likelihood parameters of \texttt{CPLL} by regarding the ground truth as a latent variable. 
In the M-step, we train a naive classifier $f$ to predict the true posterior $P(Y=y|T=t, X=\mathbf{x})$ via a true posterior estimator (Section \ref{sect:true posterior estimator}).
In the E-step, we update the confidence score via a confidence estimator (Section \ref{sect:confidence estimator}), which consists of the prior confidences (calculated from annotations) and posterior confidences (learned by model).

\subsection{True Posterior Estimator}
\label{sect:true posterior estimator}
First, we train a naive classifier as our true posterior estimator $f$ to infer the true posterior $P(Y=y|T=t, X=\mathbf{x})$. 
To model the sequence, we adopt a pre-trained language model (BERT \cite{kenton2019bert}) $\mathcal{M}$ to learn a content-aware token representation.
Specifically, we input the sequence $\mathbf{x}=\{t_1, t_2, ..., t_{|\mathbf{x}|}\}$ into $\mathcal{M}$ to obtain the sequence representations,
\begin{equation}
    H = \mathcal{M}(\mathbf{x}, \theta_{\mathcal{M}})
\end{equation}
where $\theta_{\mathcal{M}}$ is the parameters of $\mathcal{M}$, $H=[h_1, h_2, ..., h_{|\mathbf{x}|}]$. $h$ is token $t$'s content-aware representation. 

Then, we utilize a fully connected layer (FC) to predict the probability distribution,
\begin{equation}
    f(y; t, \mathbf{x}) = \sigma(W*h+b)
\end{equation}
where $\sigma$ is a sigmoid function, $\theta_{FC}= \{W, b\}$ is the learnable parameters of FC. We regard $\theta=\{\theta_{\mathcal{M}}, \theta_{FC}\}$ as a parameter set of true posterior estimator $f$. 
Negative learning \cite{kim2019nlnl} is adopted, which not only considers ``the token belongs to positive label (candidate label $y \in \hat{\mathbf{y}}$)" but also ``the token does not belong to negative label (its complementary label $y \notin \hat{\mathbf{y}}$)". 
The loss function is computed,
\begin{equation}
\small
\mathcal{L}(f(y; t, \mathbf{x}), y) = \left\{
             \begin{array}{lr}
             -\log(f(y; t, \mathbf{x})), & y \in \hat{\mathbf{y}}\\
             -\log(1-f(y; t, \mathbf{x})), & y \notin \hat{\mathbf{y}}\\
             \end{array}
\right.
\end{equation}

Finally, we optimize the empirical risk by integrating confidence $g(y; \hat{\mathbf{y}}, t, \mathbf{x})$ with the loss function (Equation \ref{eq1}). We will introduce the confidence $g(y; \hat{\mathbf{y}}, t, \mathbf{x})$ in detail below.

\subsection{Confidence Estimator}
\label{sect:confidence estimator}
The confidence estimator is used to learn the confidence scores $g(y; \hat{\mathbf{y}}, t, \mathbf{x})$, which represents the confidence of label  $y$ given the token $t$, text sequence $\mathbf{x}$, and partial label $\hat{\mathbf{y}}$. 
\begin{equation}
\label{equ:confidence}
    g(y; \hat{\mathbf{y}}, t, \mathbf{x}) = \alpha * c^A_{y; t, \mathbf{x}} + (1-\alpha)*c^M_{y; t, \mathbf{x}}
\end{equation}
where the confidence score $c^M_{y; t, \mathbf{x}}$ is learned by model and $c^A_{y; t, \mathbf{x}}$ is given by annotators. $\alpha$ is a hyper-parameter used to balance these two terms. 
The annotators will affect the quality of the datasets and we can calculate the prior confidence based on the labeled times of each class. 
However, prior confidence is biased since the annotators we selected have biases.
To address this problem, we also let the model learn the posterior confidence to reduce the biases in prior confidence. 

\paragraph{Posterior Confidence} We update posterior confidence $c^M_{y; t, \mathbf{x}}$ based on true posterior distribution $P(Y=y|T=t, X=\mathbf{x})$ estimated by true posterior estimator $f(y; t, \mathbf{x})$.
\begin{equation}
c^M_{y; t, \mathbf{x}} = \left\{
             \begin{array}{lr}
             \frac{exp(P(Y=y|T=t, X=\mathbf{x}))}{\sum_{\hat{y} \in \hat{\mathbf{y}}}{exp(P(Y=\hat{y}|T=t, X=\mathbf{x}))}}, & y \in \hat{\mathbf{y}}\\
             \frac{exp(P(Y=y|T=t, X=\mathbf{x}))}{\sum_{\hat{y} \notin \hat{\mathbf{y}}}{exp(P(Y=\hat{y}|T=t, X=\mathbf{x}))}}, & y \notin \hat{\mathbf{y}}\\
             \end{array}
\right.
\end{equation}

We calculate the confidence score for positive and negative labels independently.

\paragraph{Prior Confidence} We translate the labeled times $\mathbf{a}$ obtained from annotation into prior confidence $c^A_{y; t, \mathbf{x}}$.
\begin{equation}
c^A_{y; t, \mathbf{x}} = \left\{
             \begin{array}{lr}
             \frac{exp(a)}{\sum_{\tilde{a} \in \mathbf{a}} exp(\tilde{a})}, & y \in \hat{\mathbf{y}} \\
             0, & y \notin \hat{\mathbf{y}}\\
             \end{array}
\right.
\end{equation}

Note that both $c^M_{y; t, \mathbf{x}}$ and $c^A_{y; t, \mathbf{x}}$ are token and content dependent. 
The annotations are always affected by both the token self and the content of the token. 
Thus, we model the confidence by considering both the token and content.
Finally, we compute the final confidence score $g(y; \hat{\mathbf{y}}, t, \mathbf{x})$ via Equation \ref{equ:confidence}, which considers both biases from annotators and models.

We update the parameters $\theta$ and confidence score in the M step and E step of the EM algorithm. 
Specifically, we perform the true posterior estimator and confidence estimator iteratively.
The initialization of $c^M_{y; t, \mathbf{x}}$ is $\frac{1}{|\hat{\mathbf{y}}|}$ for $y \in \hat{\mathbf{y}}$ and $\frac{1}{|\mathcal{Y}| -|\hat{\mathbf{y}}|}$ for $y \notin \hat{\mathbf{y}}$.

\section{Experimental Setups}
In this section, we first introduce one real-world and four synthetic datasets we adopted to evaluate the performance (Section \ref{sect:datasets}). 
Then, we list the selected popular baselines to investigate the validity of our \texttt{CPLL} model (Section \ref{sect:baselines}). 
Finally, we present the implementation details and metrics to replicate the experiment\footnote{Our code is available at https://github.com/LemXiong/CPLL} easily (Section \ref{sect:Implementation Details and Metrics}).

\subsection{Datasets}
\label{sect:datasets}
\textbf{Real-World Dataset.} 
To build the real-world dataset, we ask the annotators to label the person, place, and time in the text independently. 
Each sample is assigned to three annotators with guidelines and several examples. 
To be specific, we ask three students to label 1000 samples as the training set. 
The average Kappa value among the annotators is 0.215, indicating that the crowd annotators have low agreement on identifying entities in this data. 
In order to evaluate the system performances, we create a set of the corpus with gold annotations. 
Concretely, we randomly select 881 sentences from the raw dataset and let two experts generate the gold annotations. 
Among them, we use 440 sentences as the development set and the remaining 441 as the test set. 
Table \ref{tab:real dataset statistical infomation} shows the statistical information of this dataset.

\begin{table}[t!]
\centering
\setlength{\tabcolsep}{0.8mm}{
\begin{tabular}{lllll}
\hlineB{3}
 & \#Sample & \#TIME & \#PLACE & \#PERSON \\ \hline
  Training            & 1000          & 6934                 & 958                   & 3518                   \\ 
Dev              & 440           & 955                  & 147                   & 351                    \\ 
Test             & 441           & 1015                 & 171                   & 356                    \\ 
\hlineB{3}
\end{tabular}}
\caption{The statistical information of real-world dataset. \#Sample means the number of samples in the corresponding dataset. \#TIME, \#PLACE and \#PERSON represent the number of time, place, and person entities.}
\label{tab:real dataset statistical infomation}
\end{table}

\textbf{Synthetic Datasets.} Inspired by \cite{rodrigues2014sequence}, we build synthetic datasets by adding noise on four typical NER datasets: MSRA \cite{levow2006third}, Weibo \cite{peng2015named}, Ontonotes 4.0 \cite{weischedel2011ontonotes} and Resume \cite{zhang2018chinese}. 
To simulate a real noise situation, we add noise to the original datasets using four rules: 
1) BE (Bound Error) that adds or deletes some tokens of the entity to destroy the bound (change ``room 1003" to ``(room 1003");
2) ME (Missing Error) that removes the entity from the label (``David" is not labeled);
3) CE (Category Error) that changes the category of the entity (change ``Location" to ``Organization");
4) SE (Segmentation Error) that splits the entity into two entities (change ``tomorrow at 10:00 am" to ``tomorrow" and ``at 10:00 am").
We run each rule randomly with a perturbation rate $r$, which is set as 10\% in the experiments. 
Additionally, we explore the influence of annotation inconsistency with different rates. Table \ref{table:perturb results} shows statistical information of these datasets based on token-level majority voting. We can find that a large number of entities are perturbed by our rules. For example, more than 40\% tokens labeled as entities are perturbed with a perturbation rate $r$ of 20\%.

\begin{table}[!]
\centering
\small
\setlength{\tabcolsep}{1.3mm}{
\begin{tabular}{lccllll}
\hlineB{3}
                                            & \multirow{2}{*}{\#Original} & \multirow{2}{*}{$r$} & \multicolumn{3}{c}{\#Error}    \\
                                            &                    &                              & BI & C & Percent \\ \hline
\multirow{4}{*}{Weibo} & \multirow{4}{*}{4951} & 5\%                        & 35      & 134        &  3.4\%   \\                      
                    &       & 10\%                                   & 143      & 546       & 13.9\%   \\
                   &      & 20\%                                   & 494     & 1706      & 44.4\%   \\
                   &    & 25\%                                  & 615     & 2411        & 61.0\%   \\ \hline
\multirow{4}{*}{Resume}     & \multirow{4}{*}{79014}                 & 5\%                               & 244     & 2011        & 2.8\%   \\
                                         &                      & 10\%                     & 920     & 7361        & 10.4\%   \\
                                            &              & 20\%                          & 2979    & 25408         & 35.9\%   \\
                                         &           & 25\%                                & 4145    & 37585     & 52.8\%    \\ \hline
\multirow{4}{*}{Ontonotes}         & \multirow{4}{*}{41203}                 & 5\%                                   & 295     & 1246   & 3.7\%       \\
                                    &             & 10\%                                & 978     & 4368          & 12.9\%   \\
                                       &            & 20\%                                 & 3151    & 14849          & 43.6\%    \\
                                        &           & 25\%                                  & 4420    & 20542      & 60.5\%   \\ \hline
\multirow{4}{*}{MSRA}        & \multirow{4}{*}{241809}                     & 5\%                                & 1439     & 6869          & 3.4\%    \\
                                       &              & 10\%                              & 5115    & 26343        & 13.0\%   \\
                                         &           & 20\%                               & 16729    & 86549        &  42.0\%  \\
                                         &         & 25\%                                 & 23163    & 120707       & 59.4\%   \\ \hlineB{3}
\end{tabular}}
\caption{The statistical information of synthetic datasets. \#Original means the number of the tokens labeled as an entity (not O) in the original dataset.
BI/C means the number of tokens that have a wrong BI/Category label but the right Category/BI label. 
Percent $=$ (BI+C)/\#Original.}
\label{table:perturb results}
\end{table}

\subsection{Baselines}
\label{sect:baselines}
To verify the effectiveness of our \texttt{CPLL} model, we compare it with several strong and typical baselines, which can be categorized into three groups: voting-based models, partial label learning-based models, and annotator-based models.
\begin{itemize}[leftmargin=*, align=left]
    \item \textbf{Voting-based models.} We select two voting-based models, entity-level and token-level voting models. The entity-level voting model obtains the ground truth by voting at the entity level. The token-level voting model calculates the ground truth by voting at the token level. A BERT-based sequence labeling model \cite{kenton2019bert} is trained based on the ground truth calculated by voting.
    \item \textbf{Partial label learning-based models.} We adopt two classic PLL baselines to utilize the crowd-annotated data with multiple candidate labels. PRODEN-mlp \cite{lv2020progressive} adopts a classifier-consistent risk estimator with a progressive identification method for PLL. \citet{wen2021leveraged} propose a Leveraged Weighted (LW) loss for PLL to take the partial and non-partial labels into account, which is proved to be risk consistency. It achieved state-of-the-art results on various computer version tasks. 
    We implement the models by translating the official codes to our NER task.
    \item \textbf{Annotator-based models.} After seeing researchers achieve great success in fully-supervised learning, we are easily going to think about how to gain fully-supervised data from crowd-annotated data when we use crowdsourcing. Seqcrowd \cite{nguyen2017aggregating} uses a crowd component, a Hidden Markov Model (HMM) learned by the Expectation-Maximization algorithm, to transform crowd-annotated data into fully-supervised data instead of simply voting at token-level or entity-level. When we get the ground truth calculated by this crowd component, we can adopt some efficient fully-supervised learning method to finish the corresponding task.
\end{itemize}

\begin{table*}[t!]
\centering
\resizebox{\textwidth}{!}{
\begin{tabular}{llcccccccccc}
\hlineB{3}
 & & \multicolumn{2}{c}{{Real-World}} & \multicolumn{2}{c}{{Ontonotes}} & \multicolumn{2}{c}{{Weibo}} & \multicolumn{2}{c}{{Resume}} & \multicolumn{2}{c}{{MSRA}} \\
  &                            & Dev               & Test               & Dev              & Test             & Dev               & Test               & Dev              & Test              & \multicolumn{2}{c}{Test}           \\ \hline
Ours & \texttt{CPLL}                     & \textbf{90.37}                 & \textbf{90.60}                  & \textbf{79.39}                & \textbf{81.47}                & \textbf{69.72}                 & \textbf{68.23}                  & \textbf{96.57}                & \textbf{96.07}                 &\multicolumn{2}{c}{\textbf{95.42}}               \\\hline
\multirow{2}*{Voting} & {{Token-level}}        & 89.45                & 90.40                 & 78.17               & 80.12               & 67.79                & 63.81                 &  95.81          & 95.39                & \multicolumn{2}{c}{94.68}              \\
& {{Entity-level}}       & 89.79                & 90.04                 & 78.02               & 79.30              & 65.59                & 59.34                 & 95.64               & 94.88                & \multicolumn{2}{c}{94.78}             \\\hline
\multirow{2}*{PLL} 
& PRODEN-mlp         & 87.39                & 87.90                 & 73.04               & 75.36               & 66.37                & 61.85                 & 93.90               & 94.90                 & \multicolumn{2}{c}{92.46}              \\
& {{LW loss}}           & 88.80                & 89.83                 & 79.07               & 80.45               & 69.63                & 64.26                 & 96.37               & 95.64                &\multicolumn{2}{c}{95.35}               \\
 \hline
Annotator
& Seqcrowd          & -                & -                 & 62.80               & 65.34               & 47.56                & 41.49                 & 92.73               & 93.30                & \multicolumn{2}{c}{91.90}              \\ 
\hline
Upper Bound & {{Clean data}}    & -                & -                 & 79.74               & 81.47               & 70.83                & 68.87                 & 96.64               & 96.31                &\multicolumn{2}{c}{95.53}             \\ \hlineB{3}
\end{tabular}}
\caption{The performance of our model and baselines in terms of F1. For real-world dataset, we do not report the results on clean data and Seqcrowd since we do not have ground truth for the training set.}
\label{table:main results}
\end{table*}

\begin{table*}[t!]
\centering
\resizebox{\textwidth}{!}{
\begin{tabular}{lcccccccccc}
\hlineB{3}
& \multicolumn{2}{c}{{Real-World}} & \multicolumn{2}{c}{{Ontonotes}} & \multicolumn{2}{c}{{Weibo}} & \multicolumn{2}{c}{{Resume}} & \multicolumn{2}{c}{{MSRA}} \\
                        & Dev               & Test               & Dev              & Test             & Dev               & Test               & Dev              & Test              & \multicolumn{2}{c}{Test}           \\ \hline
\texttt{CPLL}                     & \textbf{90.37}                 & \textbf{90.60}                  & \textbf{79.39}                & \textbf{81.47}                & \textbf{69.72}                & \textbf{68.23}                  & \textbf{96.57}                & \textbf{96.07}                 &\multicolumn{2}{c}{\textbf{95.42}}               \\\hline
w/o Posterior Confidence      & 89.51                & 90.08                 & 79.11               & 80.42              & 68.83                & 65.84                & 95.74               & 95.38     &\multicolumn{2}{c}{94.79}                 \\
w/o Prior Confidence    & 90.60                & 90.94                 & 79.68               & 80.87              & 70.57                & 64.90                & 96.21               & 95.70                & \multicolumn{2}{c}{95.20}            \\
w/o Both         & 86.73                & 86.32                 & 78.66               & 80.22              & 67.33                & 61.59                & 95.72               & 95.23                &\multicolumn{2}{c}{94.61}              \\
\hlineB{3}
\end{tabular}}
\caption{The performance of ablation studies.}
\label{table:ablation study}
\end{table*}

\subsection{Implementation Details and Metrics}
\label{sect:Implementation Details and Metrics}
We adopt a PyTorch \cite{AdamPaszke2019PyTorchAI} framework Transformers to implement our model based on GPU GTX TITAN X. 
Chinese-roberta-wwm-ext model \cite{YimingCui2019PreTrainingWW} \footnote{https://huggingface.co/hfl/chinese-roberta-wwm-ext/tree/main} is used for our true posterior estimator.
We utilize Adam optimizer \cite{kingma2014adam} to update our model and set different learning rates for the BERT module (0.00002) and the rest module (0.002). 
The max sequence length is 512, the batch size is 8 and the dropout rate is 0.1. 
We search the best $\alpha$ from 0.1 to 0.9 with step 0.1 using the development set. 
All the baselines use the same settings hyper-parameters mentioned in their paper.
Our source code will be available soon after this paper is accepted.

To measure the performance of the models, we adopt Macro-F1 as the metric, which is widely used for 
 NER \cite{yadav2018survey}. 
 In particular, we evaluate the performance on the span level, where the answer will be considered correct only when the entire span is matched.

\section{Experimental Results}
In this section, we conduct a series of experiments to investigate the effectiveness of the proposed \texttt{CPLL} model. 
Specifically, we compare our model with three kinds of strong baselines (Section \ref{sect:main results}) and do ablation studies to explore the influence of the key parts contained in \texttt{CPLL} (Section \ref{sect:ablation studies}).
Also, we investigate the influence of annotation inconsistency (Section \ref{sect:Influence of Annotation Inconsistency}) and hyper-parameter $\alpha$, which controls the balance of posterior confidence and prior confidence (Section \ref{sect:Influence of Hyper-parameter}).

\begin{figure*}[!t]
\vspace{-4mm}
    \centering
    \subfigure[Weibo]{\includegraphics[width=5 cm]{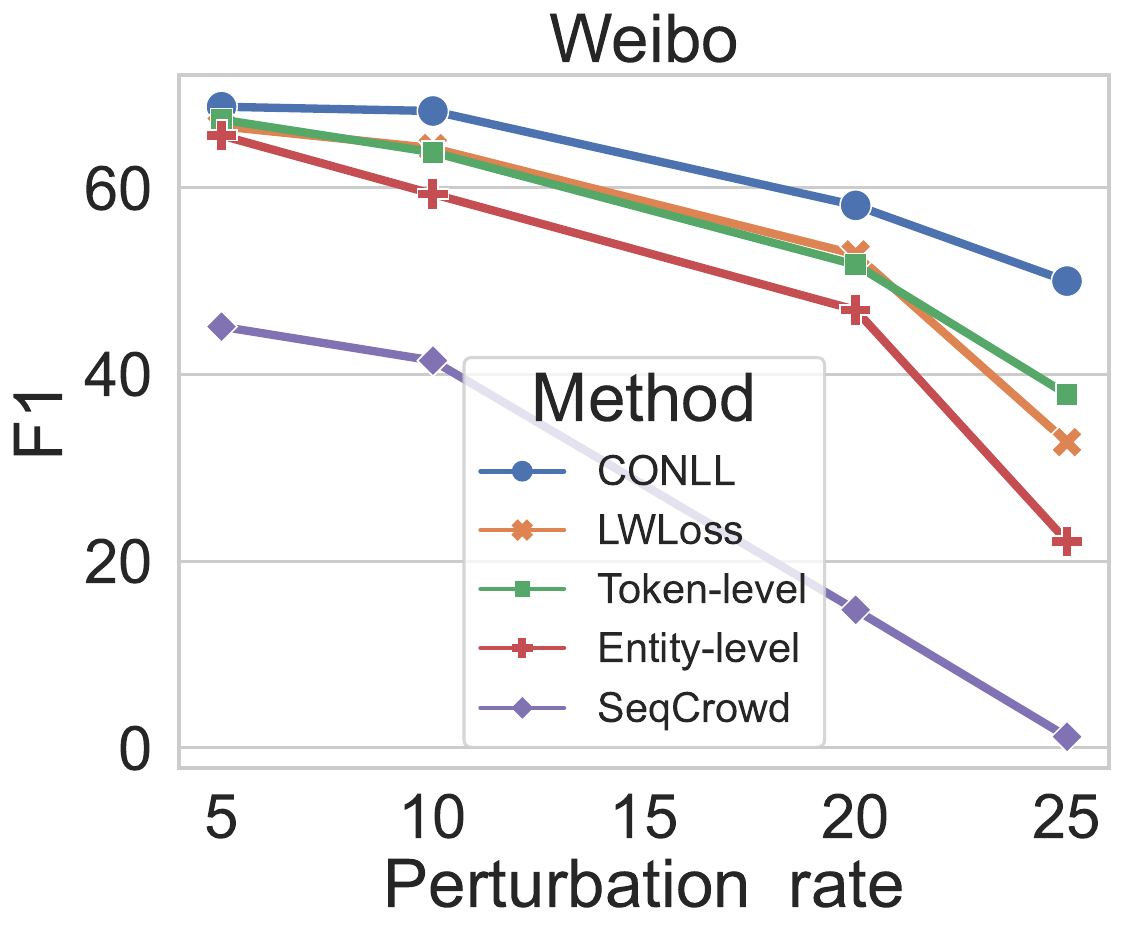}}
    \subfigure[Resume]{\includegraphics[width=5 cm]{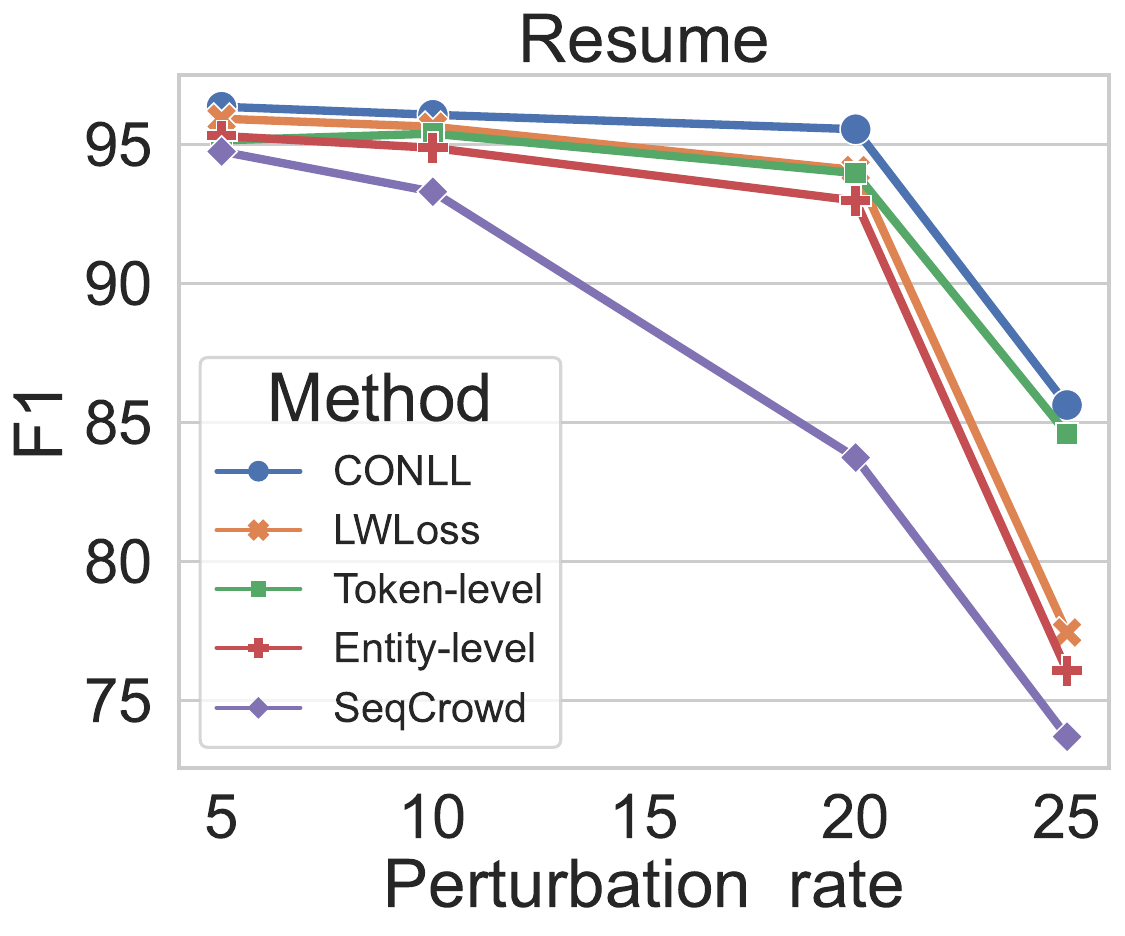}}
    \subfigure[Ontonotes]{\includegraphics[width=5 cm]{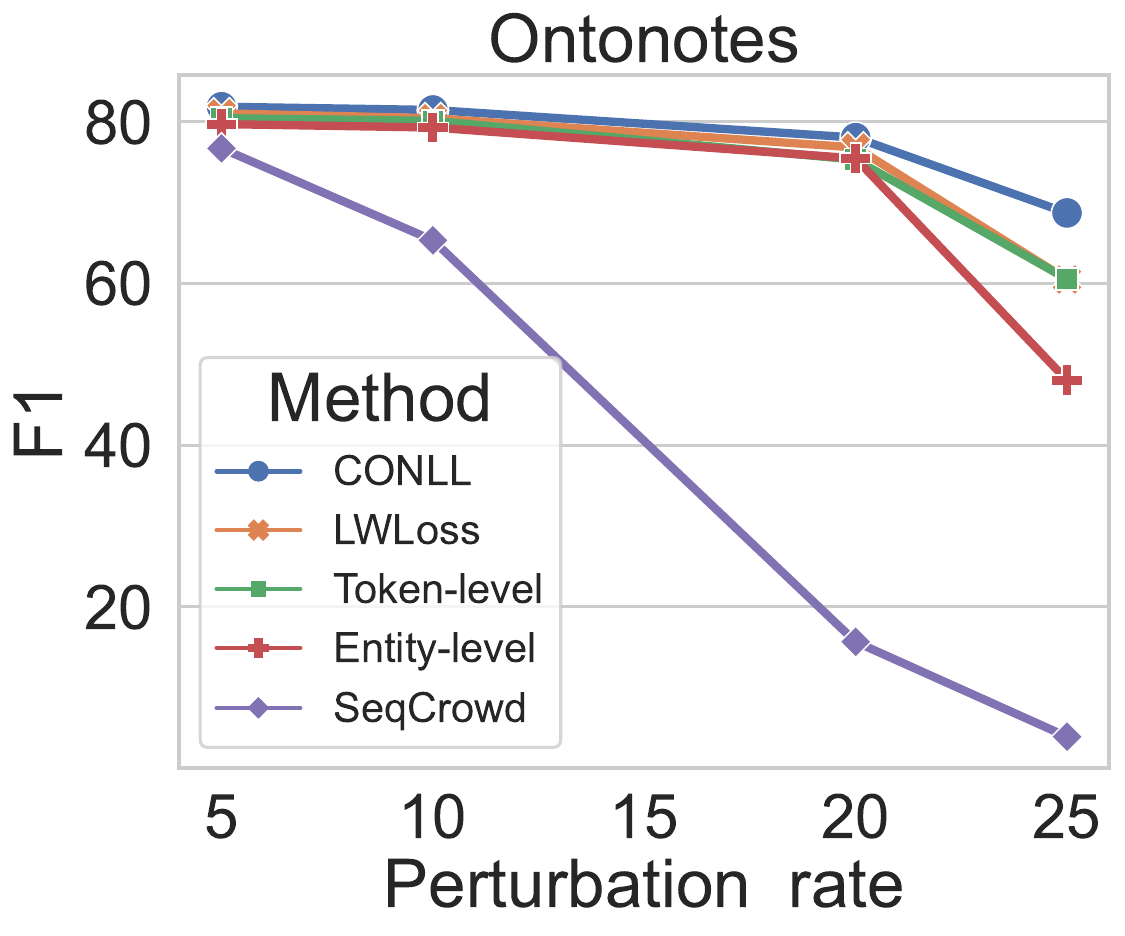}}
    \vspace{-2mm}
    \caption{The influence of annotation inconsistency.}
    \label{fig:annotation inconsistency}
    \vspace{-3mm}
\end{figure*}

\subsection{Main Results}
\label{sect:main results}
To evaluate the performance of our model, we present the results of compared baselines and our \texttt{CPLL} model (See Table \ref{table:main results}). 
\textbf{First}, we can find that our model outperforms all the baselines on both the real-world and synthetic datasets. 
The labels obtained by voting-based methods (e.g., Token-level voting and entity-level voting) always contain much noise because of the large labeling space and the complexity of this task. For PLL-based models (e.g., PRODEN-mlp and LW loss), they ignore the labeled times by the annotators. Furthermore, annotator-based methods (e.g., Seqcrowd) aim to find the trustworthy label or annotator. 
Note that Seqcrow does not work on Weibo and performs poorly on Ontonotes. It is because Seqcrow cannot solve the case of small sizes or large noise of datasets, which is also verified in Section \ref{fig:annotation inconsistency}.
All these methods cause information loss which affects the performance of the models largely.
Our \texttt{CPLL} model makes use of the crowd-annotated data by translating this task into a PLL task to integrate confidence. 
\textbf{Second}, our \texttt{CPLL} model can reduce the influence of noise effectively. From the results, we observe that \texttt{CPLL} obtains comparable results with the model trained on the clean data. Our confidence estimator can learn the bias generated by annotations effectively via the posterior and prior confidence. 

\subsection{Ablation Studies}
\label{sect:ablation studies}
To evaluate the effectiveness of each part contained in our model, we do ablation studies (See Table \ref{table:ablation study}). 
We remove posterior confidence (w/o Posterior Confidence), prior confidence (w/o Prior Confidence), and both of them (w/o Both) from \texttt{CPLL} model.
For w/o Both, we remove the confidence estimator by setting the confidences as $1/|\hat{\mathbf{y}}|$ for partial labels and $0$ for non-partial labels. 

From the results, we find the following observations.
1) Confidence estimator can learn the annotation bias effectively. Removing it (w/o Both) reduces more than 4 points in terms of F1 on the test sets over real-world and Weibo datasets.
2) Both posterior confidence and prior confidence are useful for this task. 
Obviously, prior confidence is vital to leverage the labeled confidence given by annotators. However, prior confidence may exist bias since the annotators are limited. Thus, the posterior confidence learned by the model is also crucial for partial label learning to rectify the prediction.

\subsection{Influence of Annotation Inconsistency}
\label{sect:Influence of Annotation Inconsistency}
We also explore the influence of annotation inconsistency on synthetic datasets with various perturbation rates. 
Annotation inconsistency is used to model the label quality of crowd-sourcing. 
The bigger the perturbation rate, the worse the quality of the annotation. 
We report the results with a rate from 5\% to 25\% with step 5\% over Weibo, Resume ,and Ontonotes datasets (Figure \ref{fig:annotation inconsistency}). 

First, our \texttt{CPLL} model outperforms all the baselines with different perturbation rates. 
Moreover, the higher the annotation inconsistency, the more our model improves relative to the baselines. 
Our model can reduce the influence of annotation inconsistency more effectively. 
Second, several baselines almost do not work with a large perturbation rate (e.g., 25\%), while our model can handle it effectively. 
The F1 score of Seqcrowd is only less than 20 when the rate $r$ is larger than 20\%.  
Third, it is obvious that the annotation quality will affect the performance of the model largely. The higher the inconsistency, the worse the quality of the annotation and the worse the performance of the model.

\begin{figure*}[!htb]
\vspace{-2mm}
    \centering
    \subfigure[Weibo]{\includegraphics[width=5 cm]{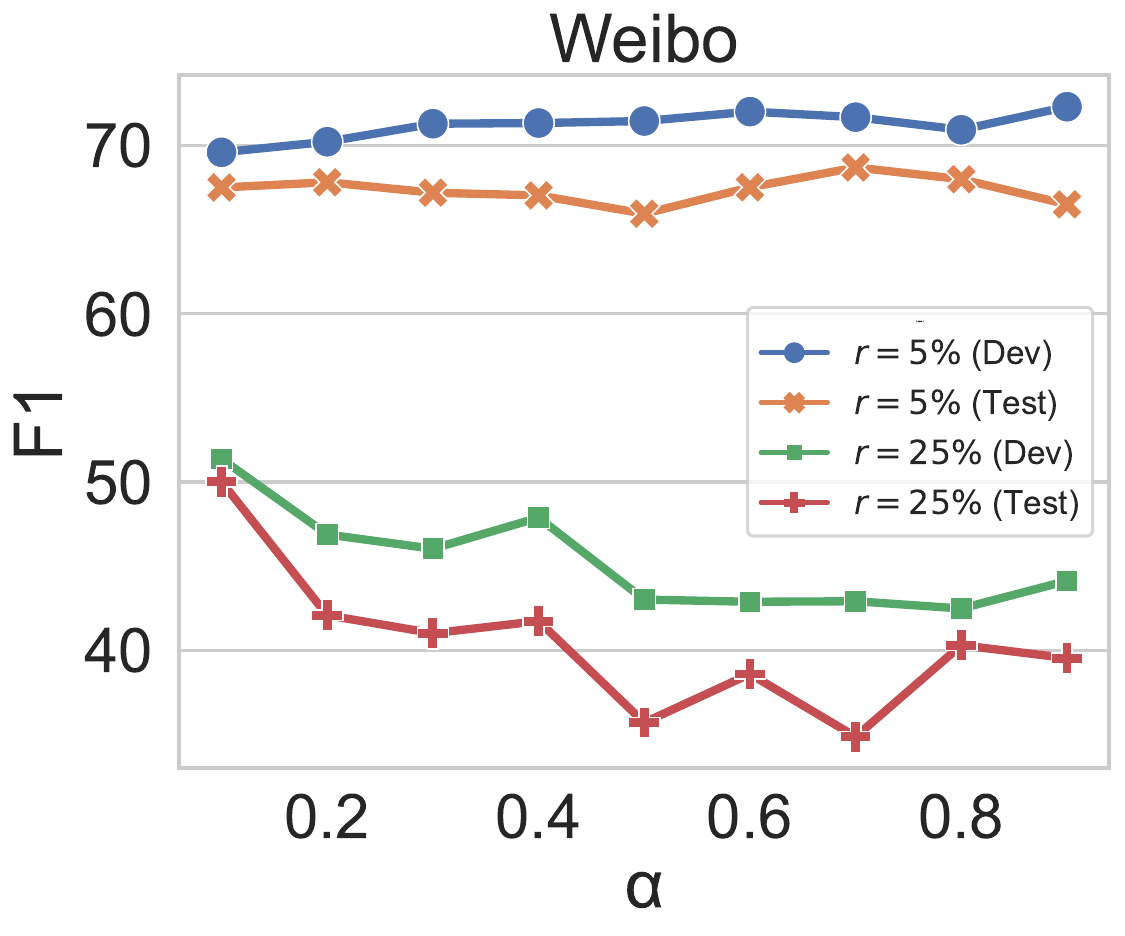}}
    \subfigure[Resume]{\includegraphics[width=5 cm]{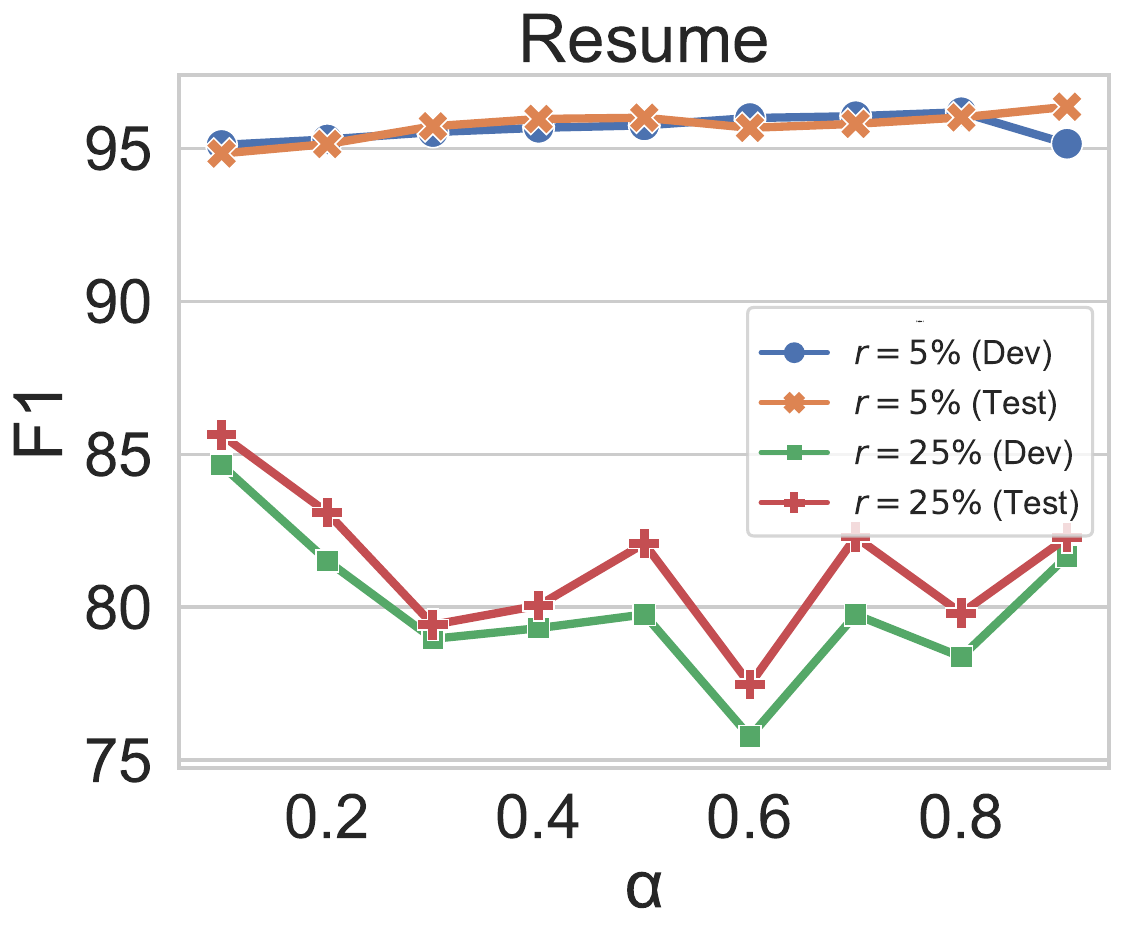}}
    \subfigure[Ontonotes]{\includegraphics[width=5 cm]{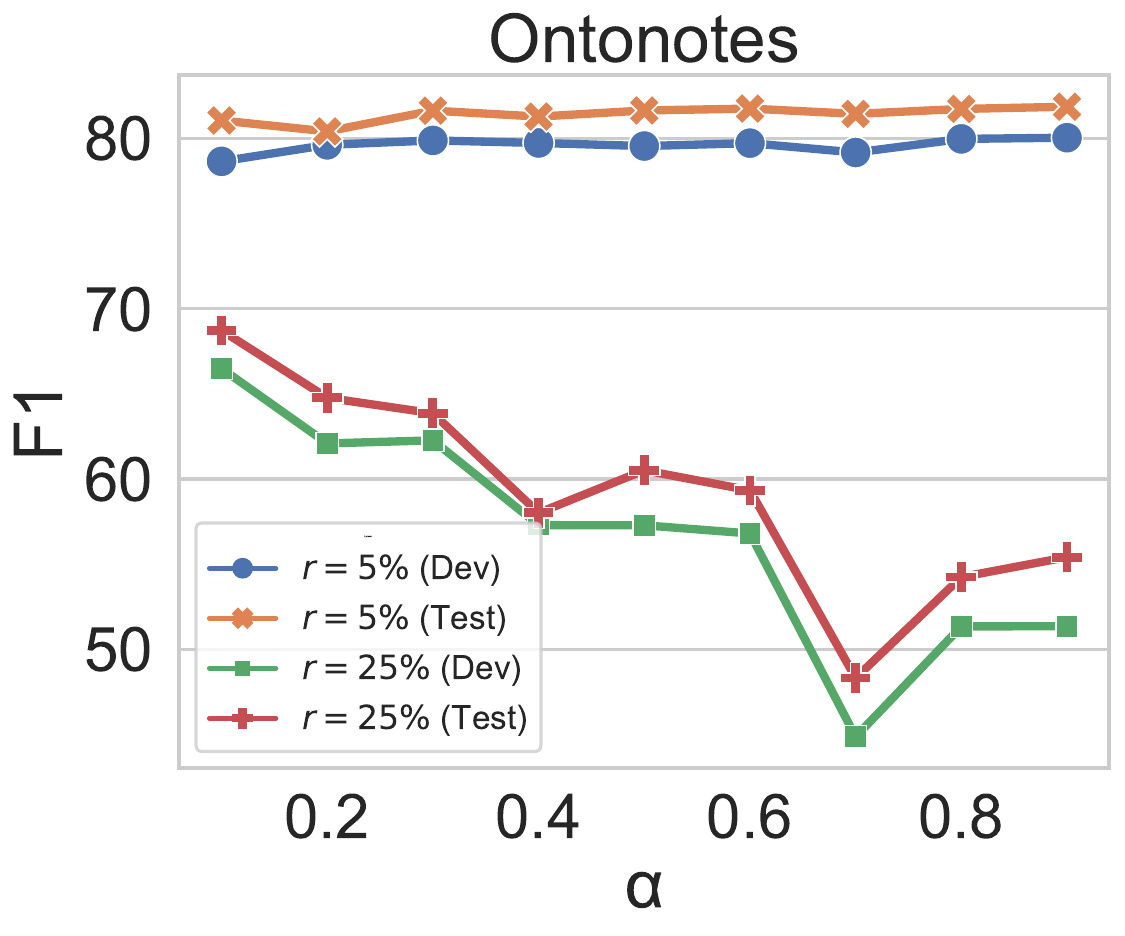}}
    \caption{The influence of hyper-parameter $\alpha$, which is leveraged to control the balance between the posterior and prior confidence.}
    \label{fig:influence of alpha}
\end{figure*}

\subsection{Influence of Hyper-parameter $\alpha$}
\label{sect:Influence of Hyper-parameter}
We further investigate the influence of the hyper-parameter $\alpha$ (in Equation \ref{equ:confidence}), which is used to balance the posterior and prior confidence (Figure \ref{fig:influence of alpha}). 
The prior confidence demonstrates the labeled confidence given by the annotators, which is biased due to the selection of annotators. To reduce this bias, we enhance our model to estimate the posterior confidence that is learned by the model. 

From the figures, we can observe the following observations. 
First, when the noise is high, the smaller the $\alpha$, the better the performance. Intuitively, the confidence given by annotators is not reliable when the perturbation rate $r$ is large. 
Second, when the noise is low, the trend that the larger the $\alpha$, the better the performance is relatively not as obvious. The reason is that the model can disambiguate the ground truth from the candidates easily since the data is clear. Most of the labels are correct and confidence is not important at this time.
All the findings indicate that our confidence estimator can make use of prior confidence and learn posterior confidence effectively. 

\section{Related Work}
In this section, we mainly review the most related works about named entity recognition (Section \ref{sect:name entity recognition}) and partial label learning (Section \ref{sect:partial label learning}).

\subsection{Named Entity Recognition}
\label{sect:name entity recognition}
Named Entity Recognition (NER) is a research hotspot since it can be applied to many downstream Natural language Processing (NLP) tasks. 
A well-trained NER model takes language sequence as input and marks out all the entities in the sequence with the correct entity type.
NER is widely treated as a sequence labeling problem, a token-level tagging task \cite{JasonPCChiu2015NamedER,AlanAkbik2018ContextualSE,HangYan2019TENERAT}. 
Also, some of the researchers regard NER as a span-level classification task \cite{MenggeXue2020CoarsetoFinePF,JinlanFu2021SpanNERNE,AlexanderAlemi2023MINERIO}. 
In these works, NER is a fully-supervised learning task based on large-scale labeled data, where each token is asserted with a golden label.

Crowdsourcing platforms (e.g., Amazon Mechanical Turk) are a popular way to obtain large labeled data.
Due to the large label space and complexity of NER, the quality of labeled data is low.
The ground truth obtained by simple majority voting contains a lot of noise, which limits the performance of the model largely.
There is some literature that trains the model from multiple annotators directly \cite{simpson-gurevych-2019-bayesian,nguyen2017aggregating}. 
They mainly focus on modeling the differences among annotators to find a trustworthy annotator. 
In fact, a sentence may not be correctly labeled by all the annotators while they all may label part of the right entities. 
To address this problem, we translate this task into a partial label learning problem with a prior confidence score.

\subsection{Partial Label Learning}
\label{sect:partial label learning}
Unlike fully-supervised learning, which uses data with golden label \textbf{y}, Partial Label Learning (PLL) asserts a candidate set $\mathcal{Y}$ for each input \textbf{x} \cite{zhang2016partial,HaoboWang2023PiCOCL,lv2020progressive}. 
Despite the fact that we can not ensure golden label \textbf{y} always in the candidate set $\mathcal{Y}$, most PLL researchers assume one of the candidate labels is the golden label for simplicity. 
The existing studies about PLL can be categorized into two groups, average-based methods \cite{zhang2015solving} and identification-based methods \cite{jin2002learning,lyu2019gm}.
Average-based methods \cite{zhang2015solving,hullermeier2006learning} intuitively treat the candidate labels with equal importance. 
The main weakness of these algorithms is that the false positive may severely distract the model with wrong label information. 
Recently, identification-based methods \cite{jin2002learning,HaoboWang2023PiCOCL} are proposed to identify the truth label from the candidates by regarding the ground truth as a latent variable. 
More and more literature pays attention to representative methods \cite{lyu2019gm,nguyen2008classification}, self-training methods \cite{wen2021leveraged}, loss function adjustments \cite{wu2018towards}. 

However, most of the current work focuses on image classification or text classification tasks, while how to model the confidence for NER is not well studied. 
The sequence labeling task aims to identify the entities in the sentence with an entity type in the token level. 
Thus, how to model the token self and its content also plays an important role in this task. 
To address this problem, we design a confidence estimator to predict the token- and content-dependent confidence based on the prior confidence given by annotators. 

\section{Conclusion and Future Work}
In this paper, we translate crowd-annotated NER into a PLL problem and propose a \texttt{CPLL} model based on an EM algorithm. 
To rectify the model's prediction, we design a confidence estimator to predict token- and content-dependent confidence by incorporating prior confidence with posterior confidence. 
We conduct the experiments on one real-world dataset and four synthetic datasets to evaluate the performance of our 
proposed \texttt{CPLL} model by comparing it with several state-of-the-art baselines. 
Moreover, we do ablation studies to verify the effectiveness of the key components and explore the influence of annotation inconsistency. 
In the future, we would like to investigate the performance of our model on other sequence labeling tasks.

\section*{Limitations}
Although our work shows that our \texttt{CPLL} model can learn from crowd-annotated NER data well, there are at least two limitations. 
First, we set the hyper-parameter $\alpha$ manually. It would be better if we could design a strategy to learn a $alpha$ adaptive value for each sample atomically. 
Second, though we mainly experiment on NER tasks, our model can be applied to all sequence labeling tasks, such as part-of-speech tagging (POS), Chinese word segmentation, and so on. We would like to explore it in further work.

\section*{Acknowledgements}
The authors wish to thank the anonymous reviewers for their helpful comments. This work was partially funded by National Natural Science Foundation of China (No.62206057), Shanghai Rising-Star Program (23QA1400200), Natural Science Foundation of Shanghai (23ZR1403500), and CCF-Tencent Open Fund.

\bibliography{anthology,custom}
\bibliographystyle{acl_natbib}




\end{document}